\documentclass{article}

\usepackage{arxiv}

\usepackage[utf8]{inputenc} 
\usepackage[T1]{fontenc}    
\usepackage{hyperref}       
\usepackage{url}            
\usepackage{booktabs}       
\usepackage{amsfonts}       
\usepackage{nicefrac}       
\usepackage{microtype}      
\usepackage{graphicx}
\usepackage{float}
\usepackage{mathtools}

\title{Distangling Biological Noise in Cellular Images with a focus on Explainability}

\author{
 Manik Sharma\\
Department of Engineering Design,\\
IIT Madras, Chennai\\
  \texttt{shmakn99@gmail.com} \\
   \And
Ganapathy Krishnamurthi\\
Department of Engineering Design\\
IIT Madras, Chennai\\
  \texttt{gankrish@smail.iitm.ac.in} \\
}

\begin{document}
\maketitle

\begin{abstract}
The cost of some drugs and medical treatments has risen in recent years that many patients are having to go without. A classification project could make researchers more efficient.

One of the more surprising reasons behind the cost is how long it takes to bring new treatments to market. Despite improvements in technology and science, research and development continues to lag. In fact, finding new treatment takes, on average, more than 10 years and costs hundreds of millions of dollars. In turn, greatly decreasing the cost of treatments can make ensure these treatments get to patients faster. This work aims at solving a part of this problem by creating a cellular image classification model which can decipher the genetic perturbations in cell (occurring naturally or artificially). Another interesting question addressed is what makes the deep-learning model decide in a particular fashion, which can further help in demystifying the mechanism of action of certain perturbations and paves a way towards the explainability of the deep-learning model.

We show the results of Grad-CAM visualizations and make a case for the significance of certain features over others. Further we discuss how these significant features are pivotal in extracting useful diagnostic information from the deep-learning model.
\end{abstract}

\section{Introduction}
It has been a human endeavour, for time eternity, to know about the inner functioning of our body, be it at the microscopic or macroscopic or macroscopic scale. In his 1859 book, On the Origin of Species, Charles Darwin propounded the earth shattering theory of Natural Selection. A reconciliation in the statistic nature and biological nature of this excitingly new phenomenon was discovered posthumously in the works of Gregor Mendel, who blamed, \emph{factors} - now called genes, as the sole reason for passage of traits or a slightly adjacent term heredity. 

Mendel's factors led scientist on frantic chase for the physical location of these genes and after around half century Alfred Hershey and Martha Chase, in the year 1952, proved definitely the seat of gene to be DNA \cite{cite1}, thought of as an useless bio-molecule till then. Followed by Har Gobind Khorana's discovery of the genetic code led to a flurry of research in the field of molecular biology that has led bare infront of scientist many exciting inner workings of the cell. This research is often accompanied by a behemoth of data, be it numerical, or be it images. A host of deep learning and machine learning techniques have been thrown at these these data-sets to tease out patterns which can be of immense practical use to the human race. 

\subsection{Biological Primer}

The process of formation of proteins, or the Central Dogma, is surprisingly not so much of a chemical concept but rather an informational one \cite{centraldogma}. The strings of coded information in the DNA provide a template for the structure of protein, but the DNA is enslaved by its geography, i.e, it is only found inside the nucleus while the protein are found in the cytoplasm (outside the nucleus). To carry this information, another molecule called RNA (specifically messenger-RNA) is employed. siRNA are single stranded molecules which bind with mRNA (the intermediary molecule in between DNA and Protein) and stops the formation of the protein \cite{sirna1}. One way to study the function of a particular gene is by silencing it and then studying the resultant cell phenotype. This is also called loss-of-function analyses of the cell \cite{knockdown}. The total silencing renders the action of a particular gene useless, which is termed as gene-knockout and it is a low through put technique as compared to gene-knockdown. Another factor to keep in mind here is the off-target effects of the siRNA used, which might affect not only the mRNA it is designed to inhibit but other mRNAs with partial matches.  

\section{Data-Set}
To understand the data fully, we need to understand the experimental setup under which the data is collected. This will help in understanding what are the sources of spurious effects which are introduced during the process of data collection (there maybe other unclassified sources as well).

\subsection{Experimental Setup}

\begin{figure}[H]
	\centering
	\includegraphics[width=0.7\linewidth]{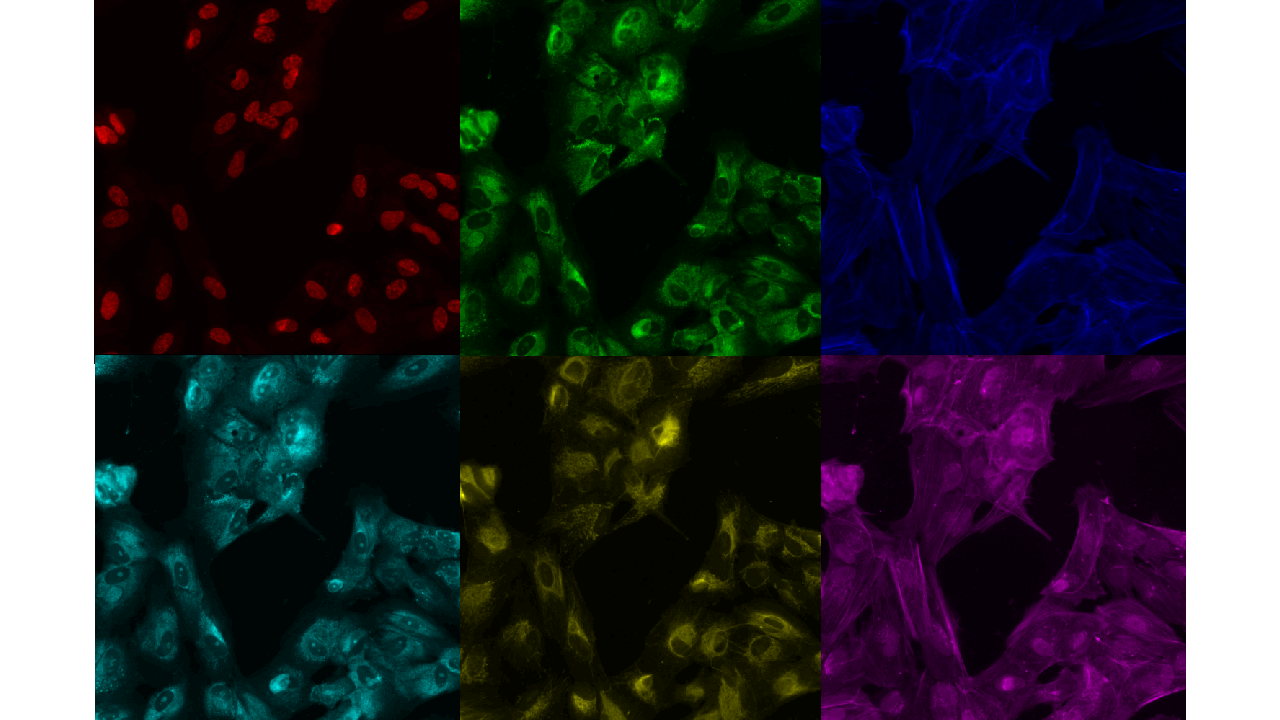}
	\caption{Channel Wise visualization of a data-point}
	\label{fg:channelvis}
\end{figure}

The experiment is conducted on single cell type, by choosing one of the four available cell types - HUVEC (Human Umbilical Vein Endothelial Cell), RPE (Retinal Pigment Epithelium), HepG2 (Human Liver Cancer Cell ) and U2OS (Human Bone Osteosarcoma Epithelial Cells). The images are taken from cell cultures. 

The siRNA is the object of interest, which is going to create the biological variability, or on a simpler level, changes which we are going to observe. Cell cultures are created in wells of plate which is can hold 384 such well ($16 \times 24$). There are in total 51 experiments conducted in the collection of database. One well can be thought as a mini-test tube. There are two imaging sites per well, images from the border wells are discarded as they might be affected by environmental noise like temperature differentials. The image obtained is of the size - $512\times512\times6$ i.e, there are six channels per image (for comparison in a standard image there are three channels - Red, Green and Blue). 

\subsection{Role of Controls}
Controls play a very crucial role in determining the performance of the model as well as calibrating the model

Negative control is a well which is left untreated in the experiments, there is one negative control per plate. A negative control is included in the experiment to distinguish reagent specific effects from non-reagent specific effects in the siRNA-treated cells \cite{knockdown}. When the image from the negative control well is given as an input the model should not make confident predictions, this is a kind of validity check

Positive control is a well which is treated with reagents whose affect on the cell culture is known and well studied \cite{knockdown}. Positive controls are used to measure how efficient the reagent is. In this context when images from the positive control are given as input the model should predict the reagent class with maximum confidence. This again serves as a sanity check. 

\subsection{Related Statistics}

There are in total $N_{Total} = 56416$ images in the data-set. Out of which $N_{Train} = 36517$ are in the train set and $N_{Test} = 19899$ are in the test set. Since the images comprise of six channels, the total number of gray-scale images in the data-set is $N_{Gray-Scale} = 338496$. There in total $N_{Class} = 1108$ classes.

\begin{figure}[H]
	\centering
	\includegraphics[width=0.5\linewidth]{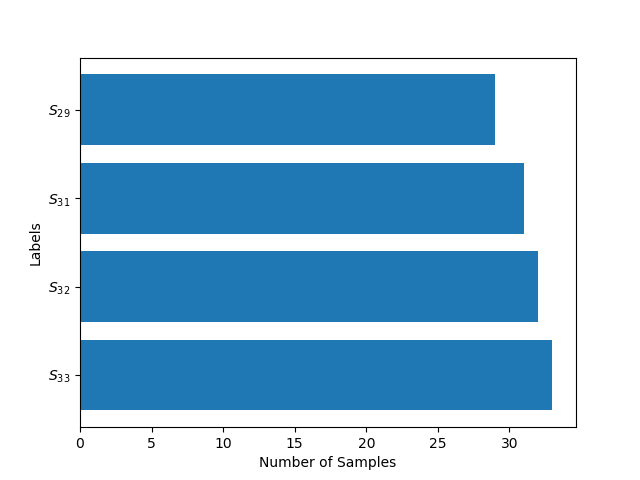}
	\caption{Class Wise Distribution of Samples}
	\label{fg:numsample}
\end{figure}

The number of samples per class, $n_{Samples/Class} = N_{Train}/N_{Class}$,  is a very important number while training the data-set. A data-set might have a high $N_{Total}$ yet a very low $n_{Samples/Class}$ in effect rendering the size of the data-set misleading. The following figure \ref{fg:numsample} shows the distribution of $n_{Samples/Class}$:

In figure \ref{fg:numsample} $S_x$ is the set of Classes which have same Number of Samples ($=x$). In Table \ref{tb:classfreq} we can see the number of classes which fall in the particular sets.

\begin{table}[H]
\begin{center}
\begin{tabular}{|l|c|}
\hline
Class Set & Number of Classes \\
\hline\hline
$S_{29}$ & $2$  \\
$S_{31}$ & $2$  \\
$S_{32}$ & $31$  \\
$S_{33}$ & $1065$  \\
\hline
\end{tabular}
\end{center}
\caption{Number of Classes in Class Sets}
\label{tb:classfreq}
\end{table}


\subsection{Pixel Value Distribution}

To get a good bearing on the distribution of pixel values in the data-set, we can look at the distribution of mean pixel value per image according to different channels.

\begin{figure}[H]
	\centering
	\includegraphics[width=0.8\linewidth]{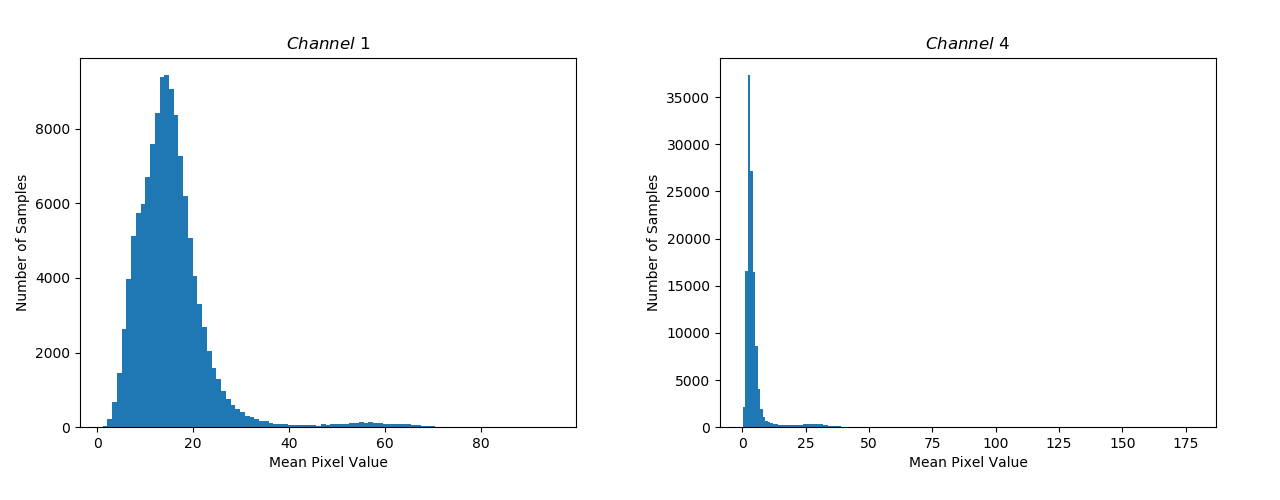}
	\caption{Channel-Wise mean Pixel Value Distribution}
	\label{fg:channelwise}
\end{figure}

The variance in the mean Pixel Values for Channel 1 is the highest and is the lowest for Channel 4 (Figure \ref{fg:channelwise}).

\section{Methods and Techniques}\label{sc:methods}

In this section we discuss techniques such as selection of backbone for the deep-learning model, the design of loss, prediction schema and methods used to understand the inner-workings of the optimized model. 

\subsection{Backbone}

Densenet takes the idea of skip connections (introduced in Resnets \cite{resnet}) one step further. In Resnet where each layer is connected to the next layer with a shortcut, in Densenet apart from the next layer each layer is connected to every other subsequent layer.

Resnets:

\begin{equation} \label{eq:resnet}
    x_l = H_l(x_{l-1}) + x_{l-1}
\end{equation}

Densenets:

\begin{equation} \label{eq:densenet}
    x_l = H_l([x_0, x_1, \ldots, x_{l-1}])
\end{equation}
where $x_l$ is the feature map after the $l^{th}$ layer. $H_l(\cdot)$ is a composite function comprising of activations, non-linearity, pooling etc. 

\begin{figure}[H]
	\centering
	\includegraphics[width=0.5\linewidth]{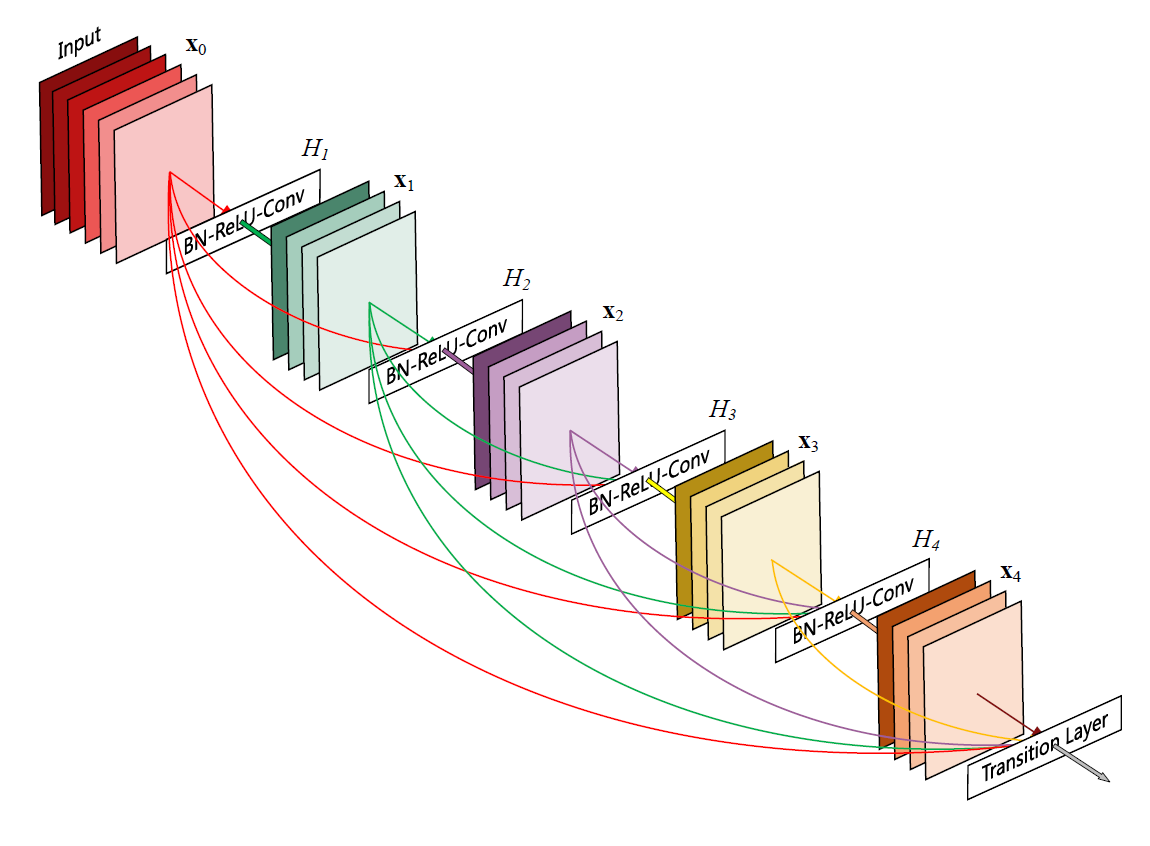}
	\caption{Architecture of a dense-block as taken from \cite{densenet}}
	\label{fg:dense}
\end{figure}

But this is all the similarity that exists between the two networks. Dense takes a departure in the way it introduces these connections, instead of summing up the input to the output, it is rather concatenated with it \cite{densenet}. In equation [\ref{eq:densenet}] the vector $[x_0, x_1, \ldots, x_{l-1}]$ is the concatenation of the feature maps outputted in layers $0,1, \ldots, l-1$.This can only be done if the dimension of the channels in the input are same as the output, therefore it is hard to maintain these skip connections through the entire network. These dense connectivity layers are packed into a module, called a dense block as shown in Figure \ref{fg:dense}. This is a miniature feed forward network in itself which has this added advantage of skip connections. These dense blocks are connected using transition layers. It is at these transition layers where the change in the dimensions of the feature map takes place. 

%

\subsection{Losses}\label{sc:losses}
Loss functions are a way to get measure of distance in the embedding space. Since the model tries to minimise this function, it encourages certain kind of structures on the embeddings produced by the model. We will trace the evolution of loss function for a particular type of problem - that of Face Recognition, because it is this class where the current task also finds its place. The difficulty which is posed by the problems of this class is that the intravariations within a class can be larger than inter-differences.

\textbf{Softmax Loss} is a go-to function which provides useful supervision signal in object detection problems. But due to the difficulty posed in the previous section, a vanilla softmax loss gives unsatisfactory results. 

\begin{equation} \label{eq:softmax}
  L_{softmax}(y,p) = -\frac{1}{N}\sum_{i=1}^{N} \log(\frac{\exp(W_{y_i}^Tx_i + b_{y_i})}{\sum_{k=1}^{K} \exp(W_k^Tx_i+b_k)})  
\end{equation}

where $N$ is the batch-size and $K$ is the number of classes.

Based on the observations of Parde~\emph{et al} \cite{cite3}, one can infer about the quality of the input image by the L2-norm of the features learnt by the softmax loss function. This gave rise to L2-softmax \cite{cite4}, which proposes to enforce all the feature to have the same L2-Norm. 

\begin{equation} \label{eq:normalize}
  \hat{W} = \frac{W}{||W||_2}, ~ \hat{x} = \alpha \frac{x}{||x||_2}  
\end{equation}

There have been other kinds of normalizations proposed for both weights and the features in the softmax loss function and it has become a common strategy in softmax. 

While the Softmax Loss did help in solving the difficulty at hand up to some extent, practitioners in the Face Recognition community wanted that the samples should be separated more strictly to avoid misclassifying the difficult samples \cite{facerec}. \textbf{Arc Loss} provides a way to deal with the misclassification of difficult samples. Arc Loss can be motivated in very straight forward way using the softmax loss function which is written as in equation [\ref{eq:softmax}], for simplicity we will assume $b_j = 0$. Writing $W_j^Tx_i = ||W_j||~||x_i|| \cos(\theta_j)$, where $\theta_j$ is the angle between the weight $W_j$ and feature $x_i$. Fixing the norm of $W_j$ to be $1$ as done in equation [\ref{eq:normalize}], following Wang~\emph{et al} \cite{cite9} we also fix the feature $x_i$ by using $l_2-normalisation$ and re-scaling it to $s$. Making the changes we get,

\begin{equation}
    L_{arc}(y,p) = -\frac{1}{N}\sum_{i=1}^{N} \\ \log(\frac{\exp(s \cos(\theta_{y_i}))}{\exp(s \cos(\theta_{y_i})) + \sum_{k=1, k\neq j}^{K} \exp(s \cos(\theta_j))})
\end{equation}

further addition of an angular margin penalty $m$ between $x_i$ and $W_j$ results in intra-class compactness and inter-class discrepancy

\begin{equation} \label{eq:arcmargin}
    L_{arc-margin}(y,p) = -\frac{1}{N}\sum_{i=1}^{N} \\ \log(\frac{\exp(s \cos(\theta_{y_i} + m))}{\exp(s \cos(\theta_{y_i}+m)) + \sum_{k=1, k\neq j}^{K} \exp(s \cos(\theta_j))})
\end{equation}

\subsection{Pseudo Labelling} \label{sc:plabel}

Pseudo Labelling is a technique borrowed from the semi-supervised learning domain. The network is trained with labeled as well unlabelled data simultaneously. In place of labels, \emph{Pseudo Labels} are used which is nothing but picking the class which has the maximum predicted probability \cite{pseudolabel}.

Let $y^{PL}$ be the pseudo label for a particular sample $x$. Let $p$ be the predicted probabilities for the sample $x$, then:

\[   
y^{PL}_i = 
     \begin{cases}
       \text{1} &\quad \text{if} ~~i = argmax_j p\\
       \text{0} &\quad \text{otherwise} \\
     \end{cases}
\]

The labelled and unlabelled samples are used simultaneously to calculate the loss value as follows:

\begin{equation}
    L = \frac{1}{N} \sum_{i=1}^{N} \sum_{k=1}^{K} loss(y_k^i, p_k^i) + \\ \alpha(t) \frac{1}{M} \sum_{j=1}^{M} \sum_{k=1}^{K} loss({y_k^{PL}}^j, p_k^j)
\end{equation}

where $p_k^j$ are the predicted probabilities for $M$ unlabelled samples and $\alpha(t)$ is a coefficient balancing the effect of labeled and unlabelled samples on the network. 

It is important to achieve a proper scheduling of $\alpha(t)$, a higher value will disrupt the training and a low value will negate any benefits which can be derived from pseudo-labelling \cite{cite8}.

One way to schedule $\alpha(t)$ is as follows:

\[   
\alpha(t) = 
     \begin{cases}
     \text{0} &\quad t ~ \le ~ T_1 \\
     \frac{t-T_1}{T_2-T_1} &\quad T_1 ~\le~t~\le~T_2 \\
     \alpha_f &\quad T_2 ~\le~ t
     \end{cases}
\]

where $T_1$ is the epoch at which the pseudo-labelling starts and increases linearly until it attains its final value at $T_2$ where it is equal to $\alpha_f$.

The reasons proposed for the success of this technique is  \emph{Low Density Separation between Classes}, according to the cluster assumption the decision boundaries between clusters lie in low density regions, pseudo labelling helps the network output to be insensitive to variations in the directions of low-dimensional manifold \cite{pseudolabel}. 

\subsection{CutMix}\label{sc:cutmix}

Data augmentation is broad variety of techniques which is used to increase the generalisation capabilities of the network, either by increasing the available train data-set or increasing the feature capturing strength of filters. There are some standard augmentation techniques such as flipping and rotating the input image which try to make the network agnostic of the pose of features in the image. Even after the application of said transforms the ability of the network to learn local features is not enhanced, there are few techniques which try to promote the object localisation capabilities.

One strategy to use is CutMix. A cropped-up part of one image is placed on top of another image and the labels are manipulated accordingly. More area one image has in the final image, the final label is higher for that image too. To achieve this first we need to make a bounding box $B$ which is the region removed from image $x_A$. Let the coordinates of the bounding box $B$ = $(r_x, r_y, r_w, r_h)$. The bounding box is sampled as follows:

\begin{equation}
    r_x \sim Uniform(0,W), ~~~ r_w = W\sqrt{1 - \lambda}\\
    r_y \sim Uniform(0,H), ~~~ r_h = H\sqrt{1 - \lambda}
\end{equation}

where $x_A, ~x_B \in \mathbb(R)^{W\times H \times C}$ and $\lambda$ is the CutMix coefficient which is sampled from a beta distribution $Beta(\alpha, \alpha)$ with $\alpha$ usually set to $1$. Also note $\frac{r_w r_h}{WH} = 1 - \lambda$ which is the cropped area ratio. Using this we get our mask $M \in \{ 0,1 \}^{W \times H}$ which is equal to $0$ for all the points inside the bounding box $B$. Therefore the final image $\hat{x}$ and label $\hat{y}$ thus become, 

\begin{equation} \label{eq:cutmix}
    \hat{x} = M \odot x_A + (1-M) \odot x_B\\
    \hat{y} = \lambda y_A + (1 - \lambda) y_B
\end{equation}

\begin{figure}[H]
	\centering
	\includegraphics[width=0.7\linewidth]{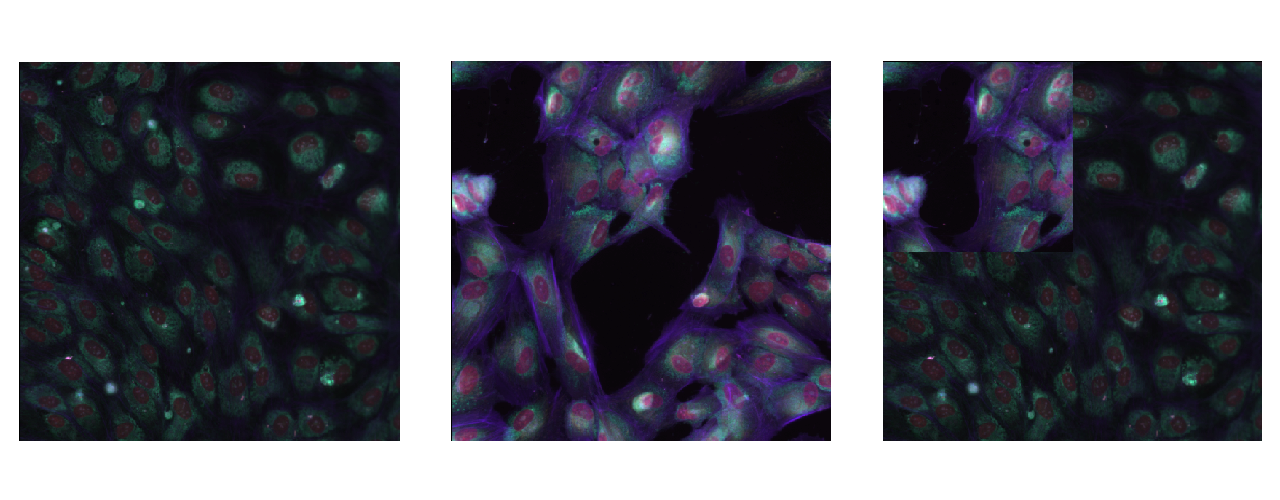}
	\caption{Image $x_A$ | Image $x_b$ | Cutmix Image}
	\label{fg:cutmix}
\end{figure}
%

\subsection{Activation Maps}\label{sc:gradcam}

There have been various techniques \cite{cite11}, \cite{cite12} proposed in recent times to understand the workings of deep-CNNs. One major break through in this has been the technique called Class Activation Maps (CAM) \cite{cam}, which proposes to produce these visualizations using global average pooling. A better and more sophisticated version has been proposed based on this work called Grad-CAM.

To go forward, we have to answer a pertinent question - What makes a good visual explanation? This can be answered in two parts. First, the technique should be able to clearly show the difference in localization properties of model for different classes, further it should show the discriminative behaviour in these localizations. Second, the visual explanations should a good resolution, therefore capturing the fine grain details about models attention. These two points are the focus of Grad-CAM.

Grad-CAM is gradient weighted global average pooling technique \cite{gradcam}, which works by first computing the importance weights $\alpha_k^c$,

\begin{equation}
    \alpha_k^c = \frac{1}{Z}\sum_{i}\sum_{j} \frac{\partial y^c}{\partial A_{ij}^k}
\end{equation}

An $\alpha_k^c$ is computed for a particular class $c$ and Activation Map $A_{ij}^k$ where the partial differentiation term $\frac{\partial y^c}{\partial A_{ij}^k}$ inside the global average pooling function is the gradient of the class activation, $y^c$(before the softmax) with respect to activation map $A_{ij}^k$

After this summation of each activation map, weighted by the importance in taken, to generate the Grad-CAM visualization $L_{Grad-CAM}^c$

\begin{equation}
    L_{Grad-CAM}^c = ReLU(\sum_{k} \alpha_k^c A^k)
\end{equation}

\section{Experiments}

\subsection{Backbone}

As compared to Resnet, a pre-trained densenet (trained on Imagenet dataset) performed better in general and was used for further training. The version used is the memory-efficient implementation of Densenet-161. The input is a six channel image of the dimensions $6\times512\times512$, the dimensions of the transforming image along with the operation acting on it are tracked in the following table:

\begin{table}[H]
\begin{center}
\begin{tabular}{|c|c|c|}
\hline
Output Size & Layer-Specification  \\[0.5ex]
\hline\hline
		$96\times256\times256$ & $7\times7$ Conv, stride = 2 \\

		$96\times128\times128$ & $3\times3$ MaxPool, stride = 2  \\

		$384\times128\times128$ &  $\left[ \begin{array}{c} 1\times1~\text{Conv},~\text{stride = 1}  \\  3\times3~\text{Conv},~\text{stride = 1} \end{array}\right]$ $\times$ 6\\

		$192\times64\times64$ &  $\left[ \begin{array}{c} 1\times1~\text{Conv},~\text{stride = 1}  \\  2\times2~\text{AvgPool},~\text{stride = 2} \end{array}\right]$ \\

		$768\times64\times64$  &  $\left[ \begin{array}{c} 1\times1~\text{Conv},~\text{stride = 1}  \\  3\times3~\text{Conv},~\text{stride = 1} \end{array}\right]$ $\times$ 12\\

		$384\times32\times32$  &  $\left[ \begin{array}{c} 1\times1~\text{Conv},~\text{stride = 1}  \\  2\times2~\text{AvgPool},~\text{stride = 2} \end{array}\right]$ \\

		$2112\times32\times32$   &  $\left[ \begin{array}{c} 1\times1~\text{Conv},~\text{stride = 1}  \\  3\times3~\text{Conv},~\text{stride = 1} \end{array}\right]$ $\times$ 36\\

		$1056\times16\times16$  &  $\left[ \begin{array}{c} 1\times1~\text{Conv},~\text{stride = 1}  \\  2\times2~\text{AvgPool},~\text{stride = 2} \end{array}\right]$ \\

		$2208\times16\times16$   &  $\left[ \begin{array}{c} 1\times1~\text{Conv},~\text{stride = 1}  \\  3\times3~\text{Conv},~\text{stride = 1} \end{array}\right]$ $\times$ 24\\
		\hline
\end{tabular}
\end{center}
\caption{Architecture of Densenet-161}
\label{tb:densenet161}
\end{table}

\subsection{Inclusion of Cell Type}

The way a siRNA reagent interacts with the cell to produce the end results, depends upto some extent on the cell-type as well. This information is incorporated in the model after the Densenet operations. There are four unique cell types, this is represented as a one-hot vector $c\in \{0,1\}^4$. The output of Densenet (a tensor of size $2208\times16\times16$) is flattened using adaptive average pooling to bring it down to size ($2208\times1\times1$) which is then further reduced in dimensions to produce a vector $output_{densenet}\in\mathbb{R}^{2208}$. Concatenating $c$ and $output_{densenet}$ we end up with a vector, $combined\in\mathbb{R}^{2212}$. 

\subsection{Loss Selection} \label{sc:losssel}

The loss function acts on a embedding, $emb\in\mathbb{R}^{1024}$ which is the output the final layer of the model. The final loss function is composite function made up of Softmax [\ref{sc:losses}] and Arc-Margin [\ref{sc:losses}] loss functions.
Using the functions defined in [\ref{eq:softmax}] and [\ref{eq:arcmargin}] we get the composite loss,

\begin{equation}
    L_{composite} = cL_{arc-margin} + (1-c)L_{softmax}
\end{equation}

The hyperparameter $s=30$ and $m=0.5$ in Equation \ref{eq:arcmargin}, the value of $c$ is $0.2$.

A composite loss function is used to produce a stable training regime. A loss function composed entirely of $L_{arc-margin}$ ended up in an oscillatory behavior. The inclusion $L_{softmax}$ helped stabilize the training process. 

\subsection{Accuracy and Loss}

\begin{figure}[H]
	\centering
	\includegraphics[width=0.7\linewidth]{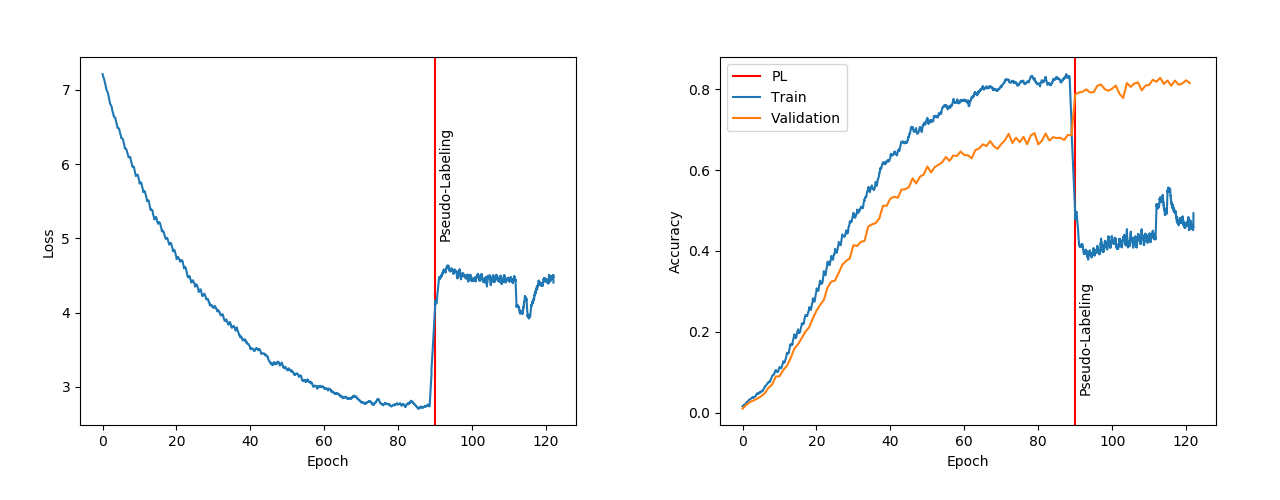}
	\caption{PL+CM+CD}
	\label{fg:accloss}
\end{figure}

As can be seen in the Figure \ref{fg:accloss} the generalising capabilities of the network were improved almost instantaneously with the introduction of Pseudo-Labelling [\ref{sc:plabel}]. 

As defined in section \ref{sc:cutmix}, CutMix helps the model generalise better. The coefficient $\lambda$ in equation [\ref{eq:cutmix}] was sampled from the distribution $Beta(\alpha, \alpha)$ with the value of $\alpha = 1$. 

Pseudo-Labelling proved better at helping the model generalise on the validation set. Though after the introduction of pseudo-labelling the training loss and training accuracy [\ref{fg:accloss}] both took a hit. We will get a better insight into the benefits of pseudo-labelling in the next section.

\subsection{Template and Feature Vector}

One way to look at the deep-learning model is through the lens of template and feature vectors. The first part of the model takes the image, $I$ and does the convolution and pooling operations, giving a feature vector $x \in \mathbb{R}^{F}$. After this a dot product is taken between $x$ and $w_i$, which is the template for the $i^{th}$ class. The resulting operation $w_i^{\top}x$ can be summarised for every class using,

\begin{equation}
    p'= Wx
\end{equation}

where $W = [w_1, w_2, \ldots ,w_C]$ and $W\in\mathbb{R}^{C\times F}$, where $C$ is the number of classes. After taking softmax of $p'$ we get $p$. The value of $p_i$ is the probability that the image $I$ belongs to class $i$. The key take away here is that, $p_i \propto w_i^{\top}x$

Therefore a model will generalise well for which $w_i$'s are spread apart. Further all the feature vectors of the same class should form a cluster and feature vectors of different classes should be far apart.

\begin{figure}[H]
	\centering
	\includegraphics[width=0.7\linewidth]{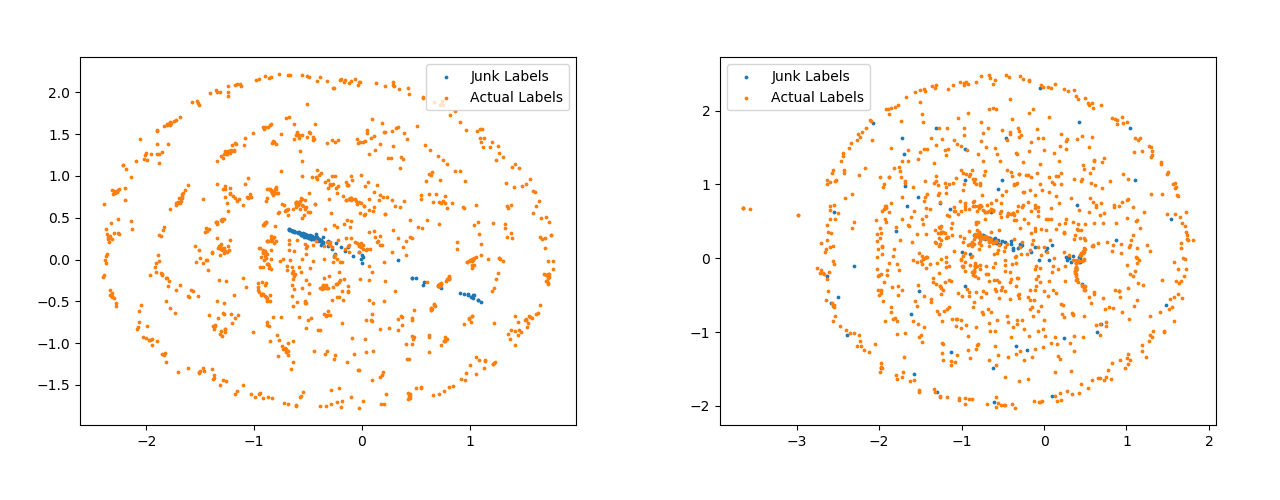}
	\caption{Arc-Margin W/o and With PL}
	\label{fg:arc}
\end{figure}

In this model we have separate template vectors for the two types of losses. The Figure \ref{fg:arc} shows the difference between the template vectors obtained obtained with and without pseudo-labelling. As compared to the vectors in left Figure \ref{fg:arc} the vectors are more spread apart in the right Figure \ref{fg:arc}.

\begin{figure}[H]
	\centering
	\includegraphics[width=0.7\linewidth]{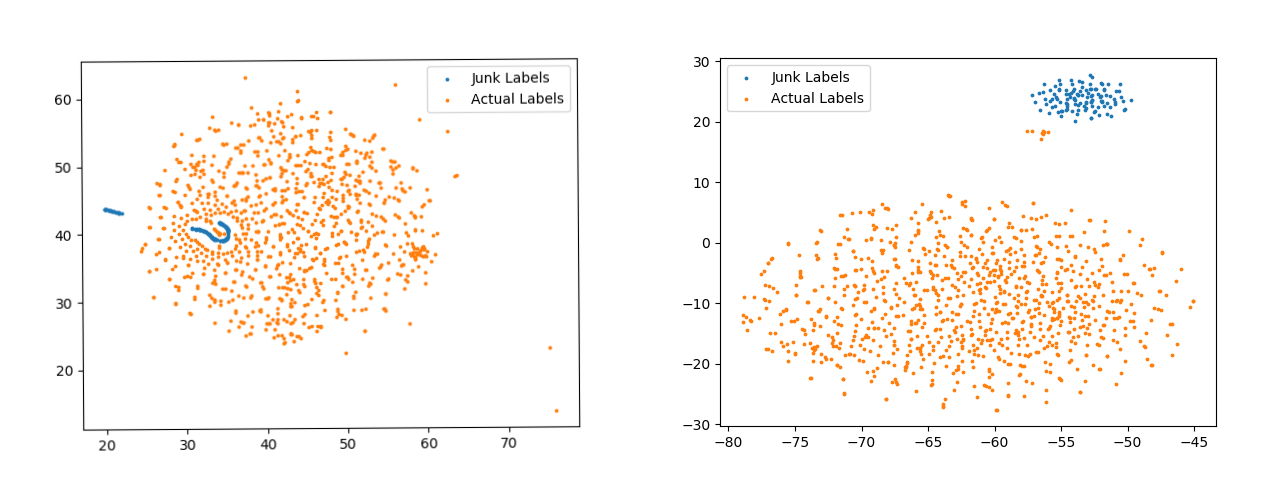}
	\caption{Softmax W/o and With PL}
	\label{fg:template}
\end{figure}

Figure \ref{fg:template} are for softmax template vectors. Apart from the spread of points, another thing to notice here are the blue points. The points shown with blue color are templates for \emph{Junk Classes}. These are the classes for which there are no samples present in the data-set. The templates generated from the model trained with pseudo-labels has made a completely separate cluster for such classes. The distinction between the Junk Classes and Actual Classes is visible all the four Figure, \ref{fg:arc}, \ref{fg:template}, but the distinction is best visible in the right Figure \ref{fg:template}.

\begin{figure}[H]
	\centering
	\includegraphics[width=0.7\linewidth]{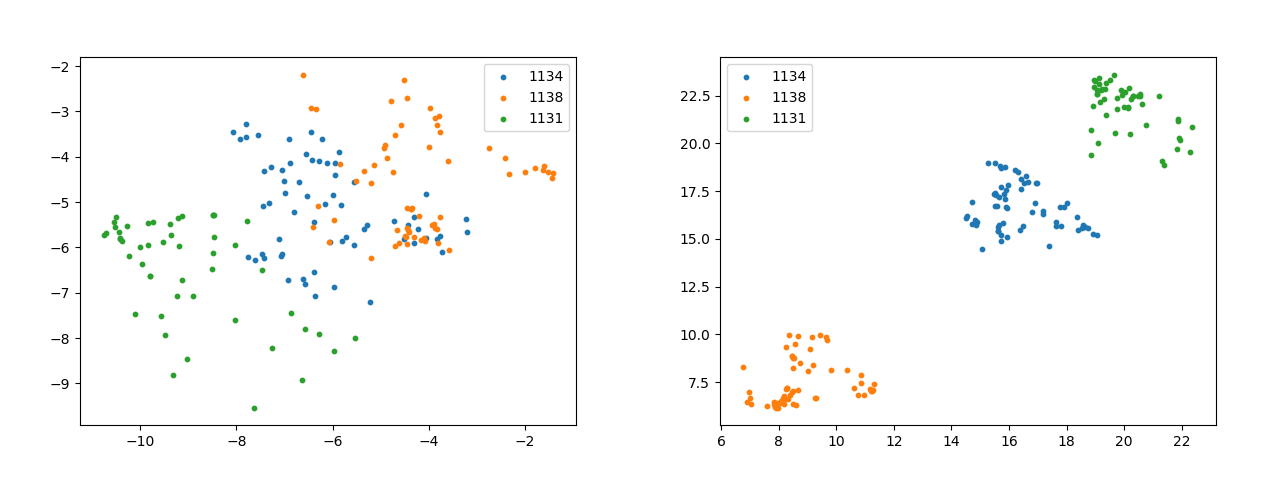}
	\caption{Features W/o and With PL}
	\label{fg:feature}
\end{figure}

Another interesting plot to look for is that of the feature vectors. The plot without pseudo-labelling (Figure \ref{fg:feature}) has vague clusters and almost in-existent decision boundary. On the other hand the clusters in Figure \ref{fg:feature} are much more well defined and tightly packed with a clear decision boundary separating the clusters. The significance further increases because of the fact that these three classes are the most prevalent in the data-set.

In the scatter Plot of class template vectors and feature vectors initially $w_i, x \in\mathbb{R}^{1024}$ which is reduced to two dimensions using t-SNE \cite{tsne}

\subsection{Convolution Layers}

Convolution layers act as filter which can be tuned in the training process to pick-up certain features and dampen others. The filter is usually square shaped and extends through all the channels of the image on which it is acting. Lets consider a square input image for simplicity (also the input image is square shaped in case). Let the input image be $I_{in}\in\mathbb{R}^{W\times W\times C_{in}}$ and the convolution filter be $f\in \mathbb{R}^{n\times n \times C_{in}}$ where $C$ is the number of input channels. The action of one filter produces a new Image with a single channel. When we club a lot of these kind of filters, we get that many images as output which can be stacked to form channels of the output image $I_{out}\in \mathbb{R}^{W' \times W' \times C_{out}}$.

\begin{figure}[H]
	\centering
	\includegraphics[width=0.7\linewidth]{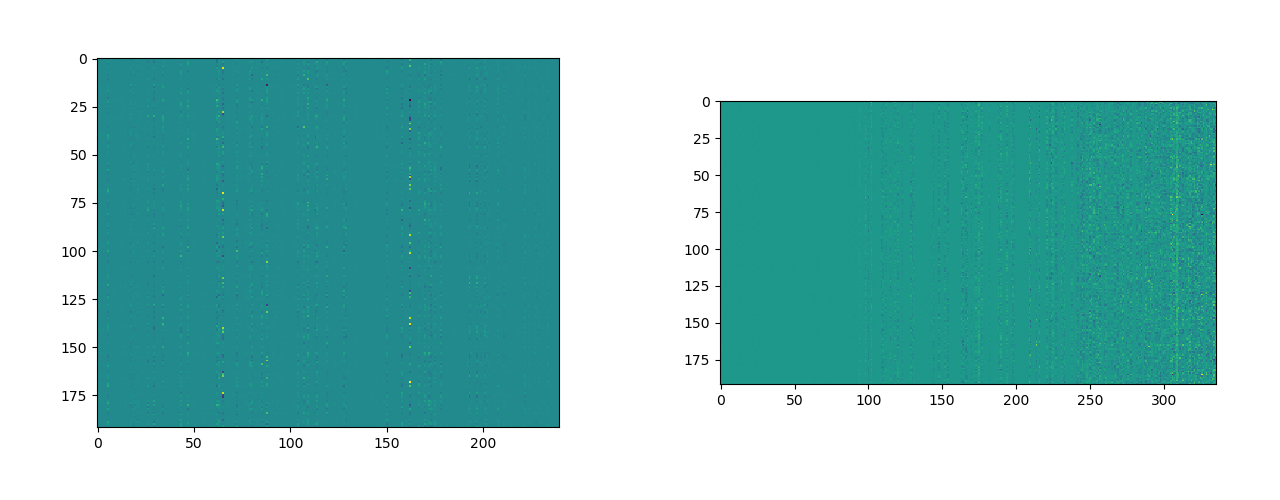}
	\caption{DB(1),DL(6),$1\times1$ and DB(2),DL(2),$1\times1$}
	\label{fg:1x1}
\end{figure}

We are interested in filters of shape $1\times 1 \times C_{in}$. The Figures in \ref{fg:1x1} shows two such filter layers. The image on the left [\ref{fg:1x1}] shows the $1\times 1$ part of the Denselayer (6) of Denseblock (1) [\ref{tb:densenet161}]. 

Another way to think of a single $1 \times 1$ convolution operation is as taking the weighted average of the channels of the input image $I_{in}$.

\begin{equation}
    I_{out,k} = w_{1,k}\times I_{in,1} + w_{2,k}\times I_{in, 2} + \ldots \\ + w_{C_{in},k} \times I_{in, C_{in}}
\end{equation}

where, $[w_{1,k}, w_{2,k}, \ldots, w_{C_{in},k}]$ is the $k^{th}$ $1 \times 1$ filter of the convolution layer, $I_{out, k}$ is the $k^{th}$ channel of the output image and $I_{in, i}$ is the $i^{th}$ channel of the input image. 

What is important to note in both the figures \ref{fg:1x1}, is the vertical bands of one-color, which are repeated horizontally, forming contiguous blocks of same color.

A band of same color signifies that instead of weighing each input channel ($I_{in, i}$) differently they multiplied by the same number, hence an average of sorts, lets call such and output $I_{avg,k}$ if this happens for the $k^{th}$ filter.

A block of same color signifies that these average images $I_{avg, k}$ are repeated multiple times in the output image. If this happens for $z$ such filters, a lot of variability in the feature map is lost, which is not a good sign.

This phenomenon is observed predominantly in two of the layers shown in Figure \ref{fg:1x1}.

%

\subsection{Grad-CAM}
Visualization techniques such as Grad-CAM [\ref{sc:gradcam}] provide us tools to better understand which part of the section of the image impacts the prediction in a positive way. This information is crucial to understand the feature selection capabilities of the network.

\begin{figure}[H]
	\centering
	\includegraphics[width=0.8\linewidth]{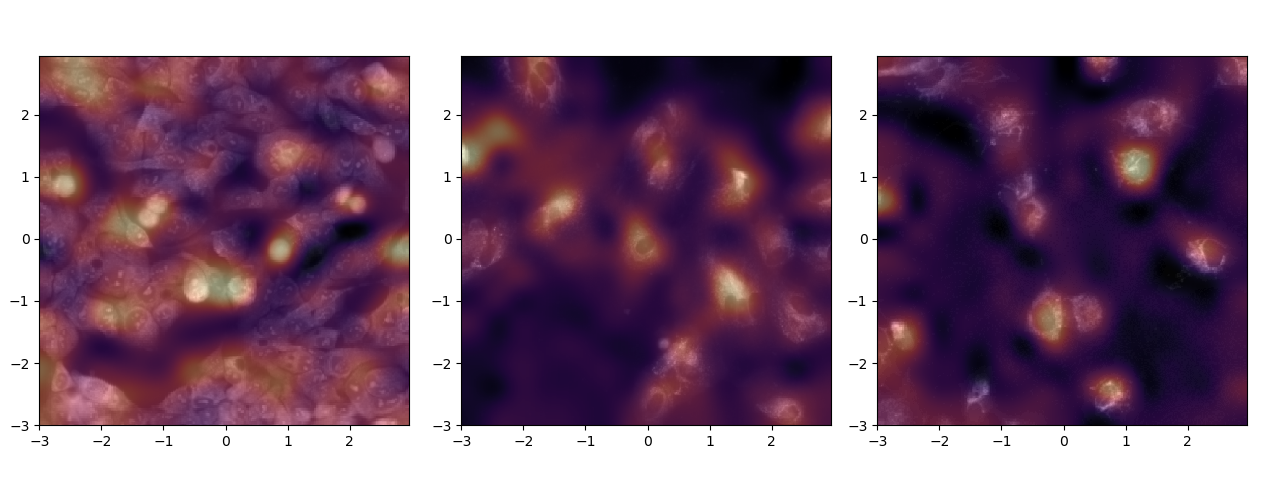}
	\caption{RPE, siRNA$_{1131}$ | HEPG2, siRNA$_{1134}$ | HEPG2, siRNA$_{1138}$}
	\label{fg:densegrad}
\end{figure}

In figure \ref{fg:densegrad}, it is quite evident that the model seem to paying attention very specific features in the image. Further it would not be wrong to say that these features are of significance, given that it is the Endoplasmic Reticulum in the left, Mitochondria in the middle and Nucleolus in the right.

\section{Discussion}
Where does a model look at while deciding the proper classification? As seen in the previous section thanks advancements in visualization techniques like Grad-CAM, this question has become answerable to a certain extent. 

A question going forward now that can be put up is  - Can we benefit from this information?  Usually when we train the model for an image classification task, the human capabilities are at par with the model (there maybe minor differences on data-sets of large scale, where sometimes the model outperforms the human or vice-versa but the magnitude is not even of one order). For problems such as these, the details about where the model pays attention serves nothing more than just a sanity check. For example a model trained to identify dogs should pay attention to the dog in the image, if a model does so we are assured that it has indeed learned to look in the right direction.

Now consider a possibility where the humans can not classify the images because there is no semantic object in the images to classify upon (which is the case in the problem at hand) and now suppose we train an agent which can do this task and is really good at it too. Wouldn't it be interesting now to look at where the agent is looking? 

\begin{figure}[H]
	\centering
	\includegraphics[width=0.3\linewidth]{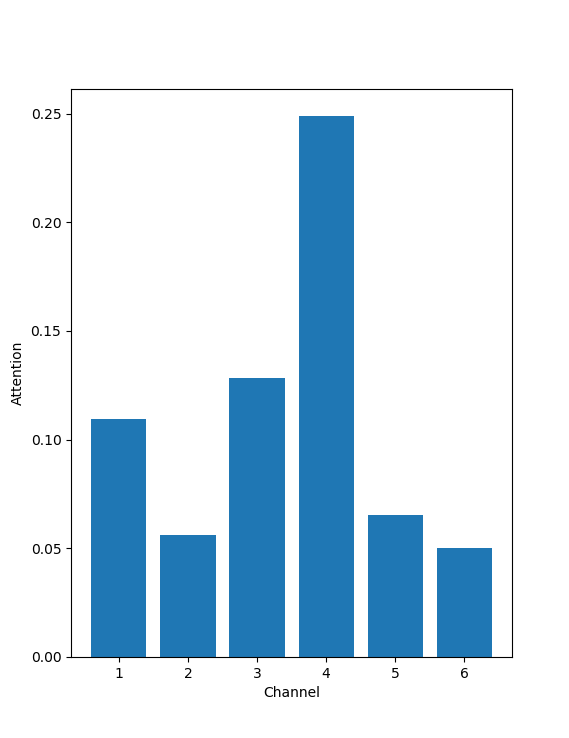}
	\caption{Plot of Attention vs. Channel for the layer - Denseblock 4, Denselayer - 24}
	\label{fg:attention2}
\end{figure}

The figure \ref{fg:attention2} is a crude representation of where the model is looking at. As can be seen most attention is payed on the Channel 5 of the image, for a given cell-type and reagent label. This should be an indication that Channel 5 is the most affected by the reagent and is the Channel which is most crucial in the prediction of the image.

Attention in this case is nothing but as follows:

\begin{equation}
    Attention_i = ||L_{CAM}(Image) - Image_i||_2
\end{equation}

Where $L_{CAM}(Image)$ is any CAM visualization and $Image_i$ is the $i^{th}$ channel of the image. This is the vector $Attention \in \mathbb{R}^C$ ($C$ is the number of channels) is then $max$-normalized to get the final attention. A higher number for a channel suggests that the attention map is concordant with channel hence the channel is significant. 

\bibliographystyle{unsrt}  
\bibliography{references}  


%
%
%

\end{document}